\documentclass[sigconf]{acmart}
\acmConference[KDD '26]{32nd ACM SIGKDD Conference on Knowledge Discovery and Data Mining}{August 9--13, 2026}{Jeju, Korea}

\acmYear{2026}
\acmISBN{}
\acmDOI{}
\startPage{1}

\setcopyright{none}
\setcopyright{cc}
\setcctype[4.0]{by}

\usepackage{booktabs}   
\usepackage{subcaption} 

\usepackage[ruled, lined, linesnumbered, commentsnumbered]{algorithm2e}
\usepackage{graphicx}
\usepackage{listings}
\usepackage{multirow}
\usepackage{subcaption} 
\usepackage{multirow}
\usepackage{soul}

\lstdefinestyle{cppstyle}{
  language=C++,
  basicstyle=\footnotesize\ttfamily, 
  numbers=left, 
  numberstyle=\tiny, 
  stepnumber=1, 
  numbersep=5pt, 
  backgroundcolor=\color{white}, 
  showspaces=false, 
  showstringspaces=false, 
  showtabs=false, 
  frame=single, 
  rulecolor=\color{black}, 
  tabsize=2, 
  captionpos=b, 
  breaklines=true, 
  breakatwhitespace=true, 
  escapeinside={\%*}{*)} 
}
\usepackage{xspace}
\usepackage{enumitem}
\usepackage{color}

\usepackage{amsmath}
\usepackage{graphicx}
\usepackage{amssymb}
\usepackage{microtype}
\usepackage{color,soul}
\usepackage{adjustbox}
\usepackage{multirow}
\usepackage[most]{tcolorbox}
\usepackage{booktabs} 

\usepackage{graphicx}
\usepackage{comment}
\usepackage{balance}
\newcommand{\sys}{SEDD\xspace}
\title{\sys: Scalable and Efficient Dataset Deduplication\\ with GPUs}

\author{Youngjun Son}
\affiliation{
  \department{Graduate School of Data Science}
  \institution{Seoul National University}
  \city{Seoul}
  \country{Korea}
}
\email{jun041577@snu.ac.kr}

\author{Chaewon Kim}
\affiliation{
  \department{Department of Computer Science}
  \institution{Seoul National University}
  \city{Seoul}
  \country{Korea}
}
\email{chaewon@aces.snu.ac.kr}

\author{Jaejin Lee}
\affiliation{
  \department{Graduate School of Data Science}
  \institution{Seoul National University}
  \city{Seoul}
  \country{Korea}
}
\email{jaejin@snu.ac.kr}

\begin{document}
\begin{abstract}
Dataset deduplication is widely recognized as a crucial preprocessing step that enhances data quality and improves the performance of large language models. A commonly used method for this process is the MinHash Locality-Sensitive Hashing (LSH) algorithm. Recently, GPU-accelerated frameworks such as NVIDIA NeMo Curator have been introduced to handle large-scale corpora; however, they remain suboptimal due to high communication overhead from physical data shuffling and underutilization of GPU resources. 
In this paper, we propose \sys, a high-performance GPU-accelerated deduplication framework optimized for distributed cluster environments. \sys introduces a computationally efficient, partially reusable hash function, alongside highly optimized GPU kernels and a hardware-aware automatic parameter selection mechanism. By replacing traditional data shuffling with a streaming-based approach, \sys significantly mitigates communication bottlenecks.

Our framework outperforms the CPU-based deduplication tool in SlimPajama by up to 158$\times$ and the GPU-based tool in NVIDIA NeMo Curator by up to 7.8$\times$ when processing 30 million documents on a node with four GPUs. Notably, \sys dramatically accelerates the previously time-consuming MinHash signature generation phase, achieving speedups of up to 375$\times$ over the CPU baseline. Despite these gains in efficiency, \sys maintains high deduplication fidelity, with duplicate document sets achieving Jaccard similarities of over 0.95 compared to those identified by the standard MinHash algorithm. In large-scale experiments, the deduplication of 1.2 trillion tokens is completed in just 3 hours on an 8-node 32-GPU V100 cluster. The related code is publicly available on GitHub (https://github.com/mcrl/SEDD).
 
\end{abstract}



\maketitle

\section{Introduction}

Pretrained language models (PLMs) have demonstrated remarkable performance across a wide range of applications~\cite{Vaswani:17, raffel2020exploring, brown2020language, touvron2023llama_a, touvron2023llama_b, dubey2024llama}, with their capabilities continuing to improve as training datasets scale~\cite{hoffmann2022training, rae2021scaling}. 
However, web-scale corpora used for modern LLM training often contain a substantial amount of duplicate and near-duplicate content~\cite{elazar2024whatsbigdata, magnusson2023paloma}. 
Such redundancy not only wastes significant computational and storage resources during training but also introduces skewed data distributions due to over-represented documents~\cite{sorscher2022beyond, lee:22}. 
Consequently, large-scale \textit{dataset deduplication} has become a critical preprocessing step in modern LLM data pipelines. 
However, scalable near-duplicate detection over trillion-token corpora remains prohibitively expensive with existing approaches.

Duplicate detection methods can be broadly categorized into two approaches: \textit{exact matching}, which identifies identical strings or hash values, and \textit{approximate matching}, which detects documents with high content similarity~\cite{albalak2024survey}. 
While exact matching is computationally efficient, it fails to capture semantically similar documents with minor variations. 
In contrast, approximate matching is significantly more computationally intensive.
A widely adopted technique for approximate matching is MinHash LSH~\cite{indyk:98}, which provides a scalable approximation of the Jaccard similarity~\cite{broder1997resemblance}. 
Despite its popularity, existing MinHash LSH implementations often struggle to efficiently process datasets at the trillion-token scale due to high computational cost and limited hardware utilization.

Recently, NVIDIA introduced NeMo-Curator~\cite{Jennings_NeMo-Curator_a_toolkit}, a GPU accelerated deduplication pipeline that significantly outperforms traditional CPU-based approaches. 
However, our analysis shows that its performance remains suboptimal in many practical environments, where excessive communication during data shuffling stalls the pipeline and results in low GPU utilization. These limitations prevent current systems from fully exploiting the massive parallelism offered by modern GPUs.

To address these challenges, we propose \sys, a high-performance GPU-based deduplication framework that maximizes end-to-end processing throughput. By combining computationally efficient hash functions with highly optimized GPU kernels and communication-aware pipeline design, \sys removes major bottlenecks in existing frameworks, including excessive data shuffling overhead and poor GPU utilization.
As a result, our system can deduplicate 1.2 trillion tokens in just 3 hours using an 8-node cluster equipped with 32 V100 GPUs.

The contributions of this paper are summarized as follows:

\begin{itemize}[leftmargin=*, itemsep=3pt]
\item \textbf{End-to-end GPU deduplication framework.} 
We present \sys, a framework that achieves 7$\times$ speedup over NeMo-Curator—the fastest existing GPU-based baseline—without compromising deduplication accuracy. 
Our design enables efficient processing of trillion-token-scale datasets on standard GPU clusters.

\item \textbf{Optimized MinHash LSH computation.} 
We accelerate MinHash generation and comparison by adopting lightweight, reusable hash functions and carefully optimized GPU kernels that improve memory efficiency and parallel utilization.

\item \textbf{Communication-efficient data pipeline.} 
We eliminate major data shuffling bottlenecks through a streaming-based execution strategy. 
Combined with double buffering, \sys reduces total communication overhead to about 20\% of that of existing frameworks.
\end{itemize}

\begin{figure*}[!ht]
\centering
\includegraphics[width=0.9\linewidth]{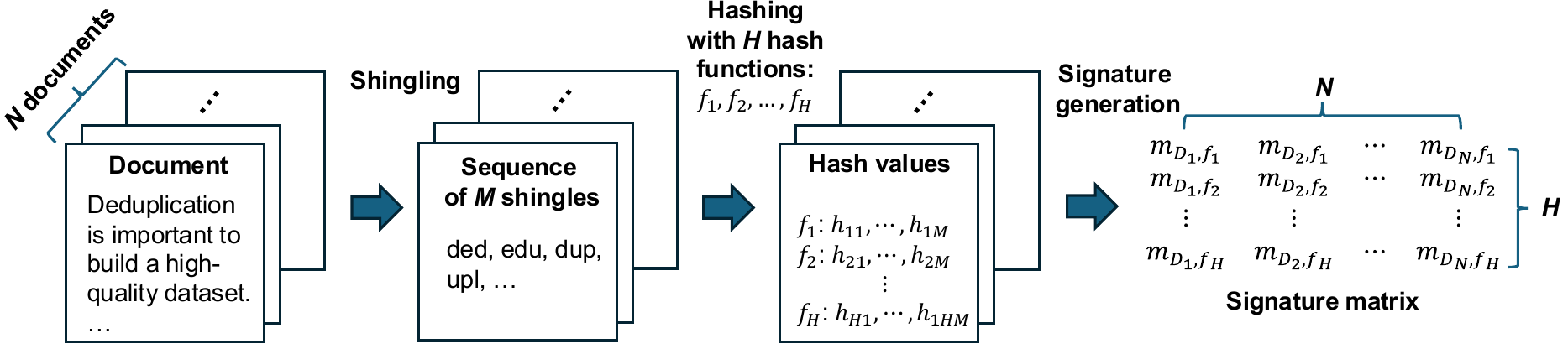}
\caption{The process of MinHash generation. }
\label{fig:minhash}
\end{figure*}

\begin{figure}[!ht]
\centering
\begin{tcolorbox}[title={RealNews Dataset}, colback=white, coltitle=black, colbacktitle=white ,boxrule=1pt, fonttitle=\bfseries, fontupper=\small, fontlower=\small]
\textbf{index:} 12347 \\
\textbf{text:} Margins matter. The more Synovis Life Technologies (Nasdaq: SYNO) keeps of each buck it earns in revenue, the more money it has to invest in growth, fund new strategic plans, or (gasp!) distribute to shareholders. Healthy margins often separate pretenders from the best stocks in the market. ... (more)
\tcblower
\textbf{index:} 39083 \\
\textbf{text:} Margins matter. The more Helen of Troy (Nasdaq: HELE) keeps of each buck it earns in revenue, the more money it has to invest in growth, fund new strategic plans, or (gasp!) distribute to shareholders. Healthy margins often separate pretenders from the best stocks in the market. ... (more)
\end{tcolorbox}
\vspace{-0.5\baselineskip}
\caption{Examples of duplicate documents.}
\label{fig:dup_example}
\end{figure}

\section{Background}
This section describes the background and related work to \sys. We examine the impact of data deduplication on the language models. Second, we describe the MinHash LSH algorithm, a standard approach for approximate duplicate detection. Finally, we introduce existing implementations, focusing on the architectural bottlenecks in distributed CPU and GPU-based frameworks that motivate our proposed design.

\subsection{Effects of Data Deduplication}

Data deduplication is the process of identifying and removing redundant entries to ensure data uniqueness, thereby improving both training efficiency and data quality for PLMs. Figure~\ref{fig:dup_example} illustrates examples of duplicate documents in the RealNews~\cite{zellers2019defending} dataset, where document structures are nearly identical. Such redundancy can hinder the effective learning of language models.

Prior studies have shown that duplicated or near-duplicate samples negatively affect model evaluation and generalization. \citet{lee:22} report that deduplication improves validation perplexity, while \citet{allamanis2019adverse} demonstrate that duplicated code samples degrade performance on code understanding tasks. Moreover, when near-duplicate documents appear in both training and test sets, evaluation metrics can be artificially inflated, leading to an overestimation of the model's capability. Removing these overlaps enables a more faithful assessment and reduces overfitting caused by memorization of repeated sequences, ultimately improving generalization~\cite{lee:22, tirumala2023d4, albalak2024survey}.

In addition to improving model quality, deduplication substantially reduces computational overhead during training. By eliminating redundant content, the model processes fewer repeated tokens, resulting in faster training and lower operational costs without sacrificing—and often improving—model performance~\cite{sorscher2022beyond}.

\subsection{MinHash}
\label{sec:minhash}
MinHash~\cite{broder1997resemblance} is a technique approximating the Jaccard similarity between two documents. 
The standard MinHash-based deduplication pipeline consists of four stages: MinHash signature generation, duplicate pair generation, construction of a union graph, and extraction of the final duplicate list.

\paragraph{\textit{\textbf{MinHash generation.}}}
As illustrated in \autoref{fig:minhash}, each document is first decomposed into a set of \textit{shingles} (or \textit{n-grams}). 
Given $H$ hash functions $f_1, \ldots, f_H$, each shingle is mapped to an integer value, producing $H$ hashed representations. 
The MinHash signature of a document is then formed by taking the minimum hash value for each function, resulting in a signature vector of length $H$. 
For $N$ documents, this process produces an $H \times N$ \textit{signature matrix}.

\paragraph{\textit{\textbf{Duplicate pair generation.}}}
Given the MinHash signatures, we estimate the similarity between documents by comparing their signatures element-wise. The similarity is approximated as the fraction of matching entries, which serves as an estimator of the Jaccard similarity. 
Document pairs whose estimated similarity exceeds a predefined threshold $\theta$ are considered near-duplicates and added to the duplicate set.

\paragraph{\textit{\textbf{Constructing a union graph.}}}
From the set of duplicate pairs, we build a union graph that groups related documents. 
Each document is represented as a node, and an edge connects two nodes if they are identified as duplicates. The connected components of this graph correspond to clusters of near-duplicate documents.
\paragraph{\textit{\textbf{Generating the final list of duplicates.}}}
For each connected component, we select a representative document (e.g., the document with the lowest index) and remove the remaining documents from the dataset. 
This procedure yields the final deduplicated corpus while preserving a single instance from each near-duplicate group.

\subsection{MinHash LSH}
\label{sec:minhashlsh}
Computing similarity between all document pairs using MinHash signatures can incur prohibitive computational cost at large scale. To address this issue, MinHash LSH (Locality-Sensitive Hashing) combines MinHash with LSH methods. 

The MinHash LSH partitions each signature vector of length \( H \) into bands and hashes each band into a bucket, resulting in a more compact set of candidate document pairs. Specifically, the signature vector is divided into \( b \) bands, each containing \( r \) rows (so that \( H = b \times r \)). Each band is then hashed into one of \( k \) buckets. Two documents are considered candidate pairs if at least one of their corresponding band hashes maps to the same bucket. This process significantly reduces the number of pairwise comparisons to a smaller subset of document pairs. 
For each candidate pair, the full signature vectors are compared. The pair is classified as a duplicate if their similarity exceeds a predefined threshold. For further details, please refer to the Appendix ~\ref{app:MinHashGen}.

MinHash LSH has become the de facto standard for document deduplication in the development of large language models~\cite{albalak2024survey,gpt3:20,gopher:21,mahabadi2025nemotron}. Although MinHash LSH provides a scalable approximation, it still imposes a significant computational and execution-time burden. Surprisingly, the system-level efficiency has been overlooked in prior research. This motivates the design of \sys, which prioritizes hardware-aware optimizations to address the growing scale of modern datasets.

\begin{figure*}[!ht]
  \centering
  \begin{minipage}{0.95\linewidth}
    \centering
    \includegraphics[width=\linewidth]{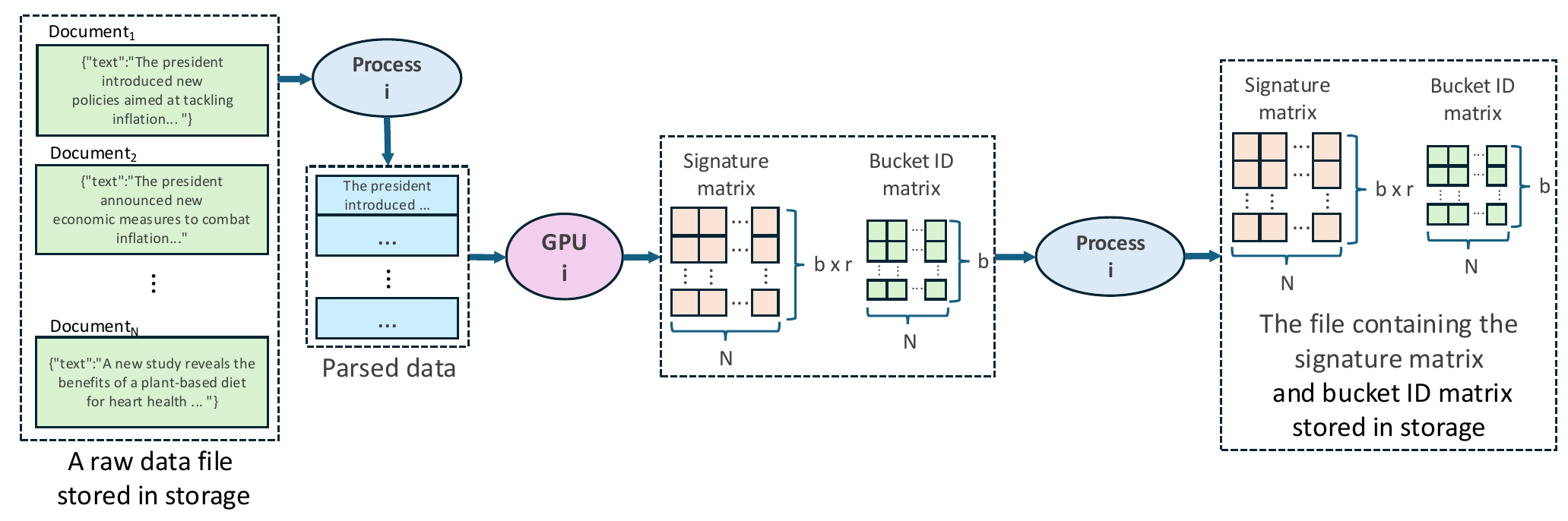}
    \vspace{-2\baselineskip}
    \subcaption{Generating the signature matrix and calculating the bucket ID for each band.}
    \label{fig:sub1}
  \end{minipage}
  \vspace{0.3cm} 
  \begin{minipage}{0.95\linewidth}
    \centering
    \includegraphics[width=\linewidth]{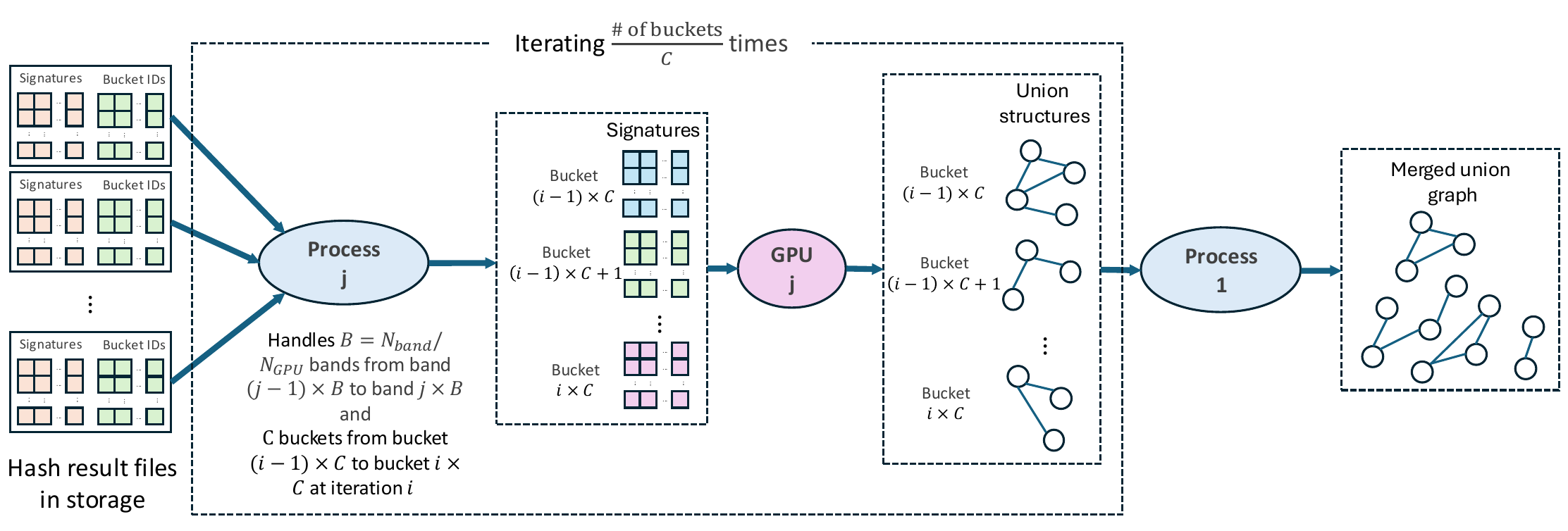}  
    \subcaption{Pairwise comparison and the union graph generation.}
    \label{fig:sub2}
  \end{minipage}
  \vspace{-1\baselineskip}
  \caption{The overview of \sys.}
  \label{fig:overall}
\end{figure*}

\subsection{Limitations of Existing Implementations}
\label{sec:existing}
\paragraph{\textit{\textbf{CPU Baseline.}}}
SlimPajama~\cite{shen2023slimpajama} employed CPU-based MinHash LSH using the Python library \textit{datasketch}. We identified implementation issues in the publicly released code and corrected them, and hereafter, we will refer to this corrected version as the \textit{CPU baseline} throughout the paper. The CPU baseline operates identically to the MinHash LSH process described in Section~\ref{sec:minhashlsh}, with one key difference. The difference lies in how comparisons are performed between documents grouped into the same bucket. When a bucket contains $D$ documents, the original approach performs pairwise comparisons among all documents, resulting in $O(D^2)$ time complexity. However, in the CPU baseline, comparisons are conducted between the first document's signature vector entering the bucket and those of subsequently arriving documents, resulting in $O(D)$. Since the CPU baseline relies on the \textit{datasketch} Python library, it is inherently inefficient for highly parallel workloads such as deduplicating billions of documents. 

Notably, search-oriented libraries such as FAISS~\cite{johnson2019billion} and ScaNN~\cite{avq_2020} are excluded from our comparison, as text deduplication for dataset curation focuses on direct token overlap and symbolic similarity rather than high-dimensional vectorization.

\paragraph{\textit{\textbf{GPU Baseline.}}}
For our primary baseline, we evaluate NVIDIA NeMo Curator~\cite{Jennings_NeMo-Curator_a_toolkit}, which currently represents the state-of-the-art in high-performance dataset deduplication. This framework provides a GPU-accelerated MinHash LSH pipeline that leverages cuDF~\cite{cudf} and Dask~\cite{dask}. While NeMo Curator offers a simplified mode that identifies duplicates based solely on bucket mapping, we specifically compare against the version that performs explicit Jaccard similarity computation. This ensures a fair and rigorous evaluation by maintaining high-fidelity standards consistent with CPU baseline and accounting for reductions in false positives.

Despite its status as a high-performance framework, NeMo Curator exhibits several architectural inefficiencies when scaling to large datasets. First, its reliance on Dask mandates a physical data shuffling phase. This process requires repartitioning the entire dataset by bucket IDs and writing intermediate results to disk, incurring substantial I/O latency and communication overhead.
Second, Nemo Curator employs a sparse bucketing strategy, using complex hash functions to map documents to a high-cardinality bucket space. Adopting the same strategy as the CPU baseline to minimize computational costs, it performs only 1-to-N comparisons within each bucket by selecting a single anchor document and comparing it against the remaining documents. Such sparse tasks may fail to provide sufficient parallel work to saturate GPU Streaming Multiprocessors, potentially leading to hardware under-utilization.
Finally, Nemo Curator follows a sequential, decoupled execution model in which each stage operates independently. This architecture requires persisting large-scale intermediate results to disk in Parquet format between stages, which increases I/O overhead. Furthermore, since each processing stage can begin only after the previous data movement completes, the framework fails to overlap communication with computation. This lack of pipeline optimization prevents the system from hiding data transfer latency, leading to a significantly longer total wall-clock time.

\section{Design and Implementation of \sys}
We develop \sys, a deduplication framework optimized for distributed GPU clusters with multiple GPUs per node. \sys is specifically designed to handle massive datasets exceeding the trillion-token scale. The end-to-end architecture is illustrated in \autoref{fig:overall}.

\subsection{Overview of \sys's Pipeline}
\label{sec:overall_process}
The deduplication process in \sys is divided into two primary phases: (1) parallel hashing and (2) streaming-based extraction with pairwise comparison.

In the first phase, raw datasets stored in formats such as JSONL are processed in parallel. Each node in the cluster spawns $N_{\text{GPU}}$ processes, where $N_{\text{GPU}}$ corresponds to the number of available GPUs. Each process manages loading document text and indices into CPU memory buffers, which are then transferred to the GPU. On the GPU, multiple CUDA threads concurrently generate MinHash signatures and assign bucket IDs for each band. These results are then returned to the CPU and persisted as intermediate hash files, maintaining a one-to-one mapping to the original input files. 

The second phase identifies and verifies duplicate candidates through a streaming-based approach. Unlike the GPU baseline, which relies on a physical shuffle to reorganize data on disk, \sys uses an on-the-fly streaming extraction strategy. Each process $j$ is assigned a specific subset of bands ($B = N_{\text{band}}/N_{\text{GPU}}$) and scans the intermediate hash result files to identify signatures belonging to the same bucket ID. To manage memory constraints and hide I/O latency, each process extracts and buffers signatures for $C$ buckets at a time. This allows pipelined execution, where the system can fetch the next batch of document data while the GPU processes the current batch. During each iteration, the extracted signatures for the $C$ buckets are dispatched to the GPU for similarity verification.

Unlike the baselines that rely on $O(D)$ anchor-based heuristics, \sys performs an exhaustive all-pairs comparison within each bucket in $O(D^{2})$, where $D$ denotes the number of documents in the bucket. Despite the theoretically higher computational complexity, \sys's hardware-optimized GPU kernels achieve superior throughput. By performing more comprehensive comparisons, \sys aims to reduce false negatives. Finally, the similarity results are aggregated to construct a global union graph for the final deduplication decision.

\subsection{Hash Functions}
\label{sec:hashfunction}
MinHash LSH performs hashing twice. The first hashing involves calculating the MinHash values for each document to generate signature matrices. The second hashing assigns bucket IDs to the bands of the signature vector of each document. The CPU and GPU baselines and \sys differ in implementing these hashing steps.

\paragraph{\textit{\textbf{Hashing in the CPU baseline}}}
For MinHash generation, the CPU baseline uses a widely used cryptographic hash function SHA-1~\cite{sha1}, which is relatively slow when processing large data. Documents are mapped to the same bucket only if their signature vectors are identical in a particular band.

\paragraph{\textit{\textbf{Hashing in the GPU Baseline}}}
Since the primary focus of MinHash generation is preventing collisions rather than cryptographic security, the GPU baseline uses the non-cryptographic MurmurHash3 algorithm~\cite{appleby2012murmurhash3} for MinHash generation. It produces a 32-bit or 128-bit hash value, enabling faster computation than SHA-1. In the LSH stage, the values of each band are mapped to buckets using the results of the MD5 hash function~\cite{rivest1992rfc1321}, which generates a 128-bit fingerprint by encoding a string of any length. 

\paragraph{\textit{\textbf{Hashing in \sys}}}
\label{sec:KEDhashing}
For MinHash generation, \sys adopts a \textit{Rabin-Karp rolling hash}\cite{karp1987efficient}, a computationally efficient non-crypto
graphic hash function that enables partial reuse across adjacent shingles. Let $s = c_{1}c_{2} \ldots c_{k}$ be a $k$-gram shingle that consists of characters. We define $f(s) = \sum_{i=1}^{k} c_{i} \cdot q^{i-1}$ such that $q$ is a constant larger than the character alphabet size. Then \sys's hash function $h(s)$ for MinHash generation is defined as follows:
\begin{equation}
\label{eq:hash}
   h(s) = f(s) \bmod p
\end{equation}
where $p$ is a sufficiently large prime number (e.g., $p=4,294,967$ in our experiments), allowing both $p$ and $q$ to be represented using 32-bit integers. Unlike SHA-1 and MurMurHash3 in the CPU and GPU baselines, the proposed hash function in \autoref{eq:hash} allows reusing the hash value of the previous shingle. For example, let $t = c_{2} \ldots c_{k}c_{k+1}$ be the next $k$-shingle of a shingle $s = c_{1}c_{2} \ldots c_{k}$. Then, $h(t)$ is computed as follows:
\begin{equation}
\label{eq:our_hash}
\begin{array}{rl}
h(t) & = f(t) \bmod p\\
& = \left(\frac{f(s) - c_{1}}{q} + c_{k+1}q^{k-1}\right) \bmod p
\end{array} 
\end{equation}

Instead of recomputing $f(t)$ from scratch, only a few arithmetic operations—a single multiplication, two additions, and one division—are required. This rolling formulation substantially reduces the computational cost compared to recomputing independent hashes of the baselines. 

The second hashing phase assigns bucket IDs. While the GPU baseline uses high-entropy hashes (e.g., MD5) to map bands into a sparse, high-cardinality space, \sys deliberately employs a \textit{dense bucketing} strategy. \sys sums the $r$ values in each band and uses the remainder of the division by the predefined constant number of buckets, $K$. 
This approach maps the sparse signature space $\mathbb{N}^{r}$ into a compact, dense space $\mathbb{N}^{K}$. The choice of $K$ is a critical trade-off. A smaller $K$ increases the collision probability, leading to larger buckets and more comparisons ($O(N^2/K)$), while a larger $K$ increases the overhead of managing fragmented buffers ($O(NK)$ in our streaming setup).
To balance the overhead of stream management against the cost of pairwise comparisons, we model the total processing time $T(K)$ as $T(K) \approx aNK + b \frac{N^2}{K}$, where $a$ and $b$ are system-dependent constants. Differentiating with respect to $K$, we derive the optimal number of buckets $K_{opt} \propto \sqrt{N}$. In practice, we set $K = 4\sqrt{N}$ based on empirical calibration, which consistently produced bucket sizes sufficient to sustain high GPU occupancy
during the comparison phase, while avoiding excessive fragmentation
overhead in the streaming pipeline.

Further discussions on the suitability of the rolling hash function and the empirical analysis regarding the effect of $K$ can be found in Appendix~\ref{app:suit_hash} and Appendix~\ref{app:ablation}, respectively.

\subsection{GPU-based Pairwise Comparison}
\label{sec:gpu_comparison}
Given a bucket containing $D$ documents, let $S \in \mathbb{Z}^{D \times H}$ be the signature matrix where $S_{i}$ represents the signature vector of the $i$-th document with length $H$. The pairwise deduplication task requires computing a binary similarity matrix $M \in \{0, 1\}^{D \times D}$, defined as:
\begin{equation}
M_{ij} = \mathbb{I}\left( \sum_{k=1}^{H} \mathbb{I}(S_{i,k} = S_{j,k}) \ge \theta \cdot H \right)
\end{equation}
where $\mathbb{I}(\cdot)$ is the indicator function and $\theta$ is the similarity threshold. To reduce redundant computation, we evaluate only the upper triangular portion of $M$. 
Since this `count-match' operation is not supported by standard BLAS(Basic Linear Algebra Subprograms) libraries, we implement a custom CUDA kernel optimized for GPU execution. 
Although the operation is not a conventional matrix multiplication, it exhibits similar memory access patterns. Thus, we adopt a GEMM-inspired tiled design to improve data locality.
Each thread is assigned to a single document pair and accumulates the number of matching hash values across the signature dimension. 
The kernel iteratively loads chunks of signature values into shared memory and reuses them across multiple comparisons, reducing global memory traffic.
After processing all tiles, document pairs whose match count exceeds the threshold are written to the output buffer and returned to the CPU for union-graph construction.

\subsection{Other Optimizations}
\label{sec:opt}
This section explains the optimization details when calculating signature matrices and generating duplicate pairs.

\paragraph{\textit{\textbf{Hardware-Aware Parameter Tuning}}}
To ensure scalability across diverse hardware configurations, \sys adopts a memory-aware execution strategy that adapts to the available CPU and GPU resources. Prior to execution, the system automatically determines several key parameters to maximize resource utilization while maintaining stable operation. Specifically, \sys decides (1) whether intermediate results are retained in memory or offloaded to storage, (2) the number of buckets processed concurrently during the streaming extraction stage, and (3) the maximum bucket size permitted for GPU-based pairwise comparison. The detailed parameter selection procedure is described in Appendix~\ref{app:auto_param_selection}.

\paragraph{\textit{\textbf{Communication-computation overlapping}}}  
To maximize performance, we implement communication-computation overlapping techniques, such as double buffering~\cite{cheng2014professional}. When the GPU performs computation, a buffer containing a batch of files is simultaneously transferred to the GPU. This overlapping strategy is applied to both the minhash generation stage and the comparison stage.

 \begin{table*}[!ht]
\centering
\caption{The deduplication time in seconds on a single node is compared across the CPU baseline, the GPU baseline, and \sys. The CPU baseline uses 64 logical CPU cores. Both the GPU baseline and \sys are evaluated using four GPUs in a single node.}
\label{tab:result1}
\vspace{-0.5\baselineskip}
\resizebox{0.85\linewidth}{!}{%
\begin{tabular}{|c|c||r|r|r|r|r|}
\hline
Method    & Dataset  & Generation & 
Comparison  & \begin{tabular}{c}
Total
\end{tabular}   &  \begin{tabular}{c}
Speedup over  \\
CPU baseline
\end{tabular} & \begin{tabular}{c}
Speedup over  \\
GPU baseline
\end{tabular} \\ \hline\hline
CPU baseline  & RealNews (100 GB) & 18,361.1    &  3001.3    &21,361.4& 1.0     & -  \\ \cline{2-7} 
(SlimPajama \cite{shen2023slimpajama})  & Sampled C4 (100 GB)       & 22,748.4    &    3911.9     &  26,660.3       & 1.0   & -    \\ \hline
GPU baseline & RealNews (100 GB) & 409.5    & 639.2 & 1048.8 & 20.4 &  1.0   \\ \cline{2-7} 
(NVIDIA NeMo Curator \cite{Jennings_NeMo-Curator_a_toolkit}) & Sampled C4 (100 GB)      & 480.8     & 954.7  & 1,435.5 &      18.6    & 1.0 \\ \hline
\multirow{2}{*}{\sys}         & RealNews (100 GB) & 49.0      & 86.2  & 135.2  & \textbf{158.0} & \textbf{7.8}\\ \cline{2-7} 
& Sampled C4 (100 GB)      & 58.9      &   137.1  & 196.0  & \textbf{136.0}  & \textbf{7.3}\\ \hline
\end{tabular}
}
\end{table*}

\section{Experiment}
This section compares \sys against the existing CPU and GPU implementations of MinHash LSH. We first compare the computation speed and accuracy for deduplication. Then, we provide a detailed analysis of the architectural factors contributing to this efficiency. 

\subsection{Evaluation Environment}
\paragraph{\textit{\textbf{Target system configuration}}}
We use a 8-node GPU cluster with a storage node for our experiments. Each node is equipped with four NVIDIA Tesla V100 GPUs, with each GPU having 32GB of memory.  The detailed target system configuration is in Appendix \ref{app:system_config}.

\begin{table*}[!t]
    \centering
    \begin{minipage}{0.48\linewidth}
        \centering
        \caption{Comparing the deduplication accuracy of the MinHash LSH-based approaches and the standard MinHash algorithm.}
        \label{tab:result2}
        \vspace{-0.5\baselineskip}
        \resizebox{\linewidth}{!}{%
            \begin{tabular}{|c||c|r||c|r|}
            \hline
            & \multicolumn{2}{|c||}{0.1M documents} & \multicolumn{2}{|c|}{1M documents} \\ \cline{2-5}
            & Ratio & Jaccard & Ratio & Jaccard \\ \hline \hline
            Standard MinHash & 7,317 / 0.1M & - & 289,229 / 1M & - \\ \hline
            CPU baseline & 7,284 / 0.1M & 0.995 & 288,683 / 1M & 0.998 \\ \hline
            GPU baseline & 7,158 / 0.1M & 0.952 & 283,724 / 1M & 0.966 \\ \hline
            \sys & 7,176 / 0.1M & 0.956 & 283,264 / 1M & 0.966 \\ \hline
            \end{tabular}
        }
    \end{minipage}
    \hfill 
    \begin{minipage}{0.48\linewidth}
        \centering
        \caption{Comparison of the set of all near duplicates between NeMo and \sys.}
        \label{tab:compare_with_nemo}
        \vspace{-0.5\baselineskip}
        \resizebox{\linewidth}{!}{%
            \begin{tabular}{|l|r|r|r|r|}
            \hline
            Dataset & \#Documents & GPU baseline & \sys & Overlap \\ \hline \hline
            RealNews & 30M & 5,426,170 & 5,426,172 & 5,329,713 (98.2\%) \\ \hline
            Sampled C4 & 50M & 2,526,120 & 2,766,436 & 2,486,835 (98.4\%) \\ \hline
            Full C4 & 365M & 25,307,187 & 27,207,144 & 24,533,637 (97.0\%) \\ \hline
            \end{tabular}
        }
    \end{minipage}
\end{table*}

\paragraph{\textit{\textbf{Comparison baselines}}}
To ensure a fair and rigorous evaluation, we align the algorithmic hyperparameters for MinHash LSH across all implementations. We adopt the standard configuration of the CPU baseline, utilizing 128 hash functions partitioned into $b=16$ bands and $r=8$ rows. Following \ \citet{lee:22}, we generate shingles using 5-grams and set the Jaccard similarity threshold to 0.8. All implementations are tuned to their best-performing configurations under identical hardware constraints.

\paragraph{\textit{\textbf{Datasets used}}}
The datasets we used in our experiments are the RealNews dataset~\cite{zellers2019defending} and C4~\cite{raffel:20}. The RealNews dataset is a large English corpus of news articles from Common Crawl, and C4 is a filtered version of Common Crawl. We select these datasets for two primary reasons. First, their significant scale—comprising over 30 million documents—is essential for evaluating the computational limits of traditional MinHash LSH and for quantifying the acceleration provided by \sys. Second, as established in prior research~\cite{lee:22}, deduplicating these specific corpora has been shown to improve the downstream performance of language models. Since this effectiveness is well-documented, we focus on the efficiency of the deduplication process itself, while including supplementary experiments to verify the consistency of these effects.

Following  SlimPajama~\cite{shen2023slimpajama}, we preprocess the datasets before deduplication. We apply NFC normalization to remove non-Unicode characters, ensuring that a letter followed by a combining character becomes a single combined character. We also filter out documents with less than 200 characters.

\subsection{Processing Speed}
We divide the overall MinHash LSH process into two main phases: the \textit{Generation} phase, which includes MinHash generation and bucket mapping, and the \textit{Comparison} phase, which includes pairwise comparisons and merging.

We used the complete 100GB RealNews dataset and the sampled C4 dataset, which consists of 100GB randomly sampled from the full 750GB C4 dataset, for this experiment. 
The CPU baseline processing time is measured using Python's multiprocessing module across 64 logical CPU cores, while both the GPU baseline and \sys are measured using four V100 GPUs in a single node. The results are summarized in \autoref{tab:result1}. For deduplication on the RealNews dataset, \sys is about 158 times faster than the CPU baseline and about 7.8 times faster than the GPU baseline. Notably the MinHash generation phase alone is roughly 375 times faster than the CPU baseline. Similarly, on the C4 dataset, \sys achieves a speedup of 136 over the CPU baseline and about 7.3 over the GPU baseline. As described in Section~\ref{sec:KEDhashing}, \sys uses $K = 4\sqrt{N}$ buckets and processes $C$ buckets concurrently during the streaming stage. Further analysis of these hyperparameters and the ablation results are reported in Appendix~\ref{app:ablation}.


\subsection{Deduplication Accuracy}
\label{sec:accuracy}
Evaluating the accuracy of deduplication on web-scale corpora is inherently challenging due to the absence of labeled ground truth. To address this, we adopt different strategies depending on the size of dataset: we utilize brute-force MinHash as a pseudo-ground truth for small datasets and employ multiple proxy evaluations—including consistency checks with GPU baseline and downstream model performance for large-scale datasets.

\paragraph{\textit{\textbf{Small Dataset.}}}
For datasets of manageable size (0.1M and 1M documents sampled from RealNews), we treat the result of a brute-force all-pairs MinHash comparison (without LSH approximation) as the oracle, or \textit{Exact MinHash}. To ensure the sampled subsets contain a statistically significant number of duplicates for evaluation, we employ stratified sampling rather than random selection, preserving the cluster structures of the original corpus.

We define \textit{the set of near-duplicates} as the collection of all document pairs identified as similar. We measure the fidelity of \sys by calculating the Jaccard similarity between the set retrieved by \sys and the set retrieved by Exact MinHash. As shown in \autoref{tab:result2}, \sys achieves a set similarity score exceeding 0.95 relative to Exact MinHash. Specifically, on the 1M dataset, \sys identified 283,264 documents as duplicates, of which 281,355 overlapped with the ground truth. This high alignment confirms that \sys captures nearly all duplicate candidates found by the computationally expensive exact method, effectively minimizing false negatives despite its optimized hashing and bucketing strategy.

\paragraph{\textit{\textbf{Large Datasets.}}}
For datasets exceeding 30 million documents, running Exact MinHash becomes computationally impractical. Consequently, we demonstrate the accuracy of \sys through two indirect evaluation methods. First, we evaluate the consistency of \sys by measuring its overlap with NeMo Curator, the current state-of-the-art GPU baseline. As detailed in \autoref{tab:compare_with_nemo}, \sys successfully identifies over 97\% of the near-duplicates found by NeMo while uncovering additional instances. 
Second, we utilize downstream language model performance as a proxy for data quality. We trained LLaMA 3.2-1B\cite{dubey2024llama} on three dataset variations: (1) non-deduplicated raw data, (2) data deduplicated by NeMo, and (3) data deduplicated by \sys. Detailed configurations are in Appendix~\ref{app:model_config}. As summarized in \autoref{tab:downstream_results}, the model trained on the \sys-deduplicated dataset performs as well as the model trained on the non-deduplicated version, despite using 10\% fewer tokens. Crucially, its performance is comparable to the NeMo-deduplicated baseline across all downstream tasks. This consistency in model performance serves as strong evidence that \sys efficiently identifies and removes redundant data without compromising the linguistic integrity or the information density of the training corpus.

\begin{table}[!t]
\centering
\begin{minipage}{0.9\linewidth}
\centering
    \caption{Execution time of the MinHash generation kernel (in seconds) for different hash functions.}
    \label{tab:result6a}
    \vspace{-0.5\baselineskip}
\resizebox{0.9\linewidth}{!}{%
  \begin{tabular}{|c|c|c||c|}
    \hline
    \# of Documents & MurmurHash3 & \sys & Speedup \\ \hline \hline
    0.1M & 0.071 & 0.025 & 2.88 \\ \hline
    1M & 0.215 & 0.08 & 2.59 \\ \hline
    30M & 13.82 & 2.85 & 4.85 \\ \hline
    50M & 11.61 & 3.63 & 3.20 \\ \hline
    365M & 98.70 & 23.98 & 4.12 \\ \hline
  \end{tabular}
  }
\end{minipage}
\end{table}

\begin{table}[t]
\centering
\begin{minipage}{0.9\linewidth}
    \centering
    \caption{Downstream task performance.}
    \label{tab:downstream_results}
    \vspace{-0.5\baselineskip}
    \resizebox{0.75\linewidth}{!}{%
    \begin{tabular}{c|rrr|}
\cline{2-4}
\multicolumn{1}{l|}{}                                                                  & \multicolumn{3}{l|}{Deduplication Method}                                              \\ \hline
\multicolumn{1}{|c|}{Task}                                                             & \multicolumn{1}{r|}{\textbf{None}} & \multicolumn{1}{r|}{\textbf{NeMo}} & \textbf{\sys} \\ \hline
\multicolumn{1}{|c|}{HellaSwag~\cite{zellers2019hellaswag}}      & \multicolumn{1}{r|}{43.8}          & \multicolumn{1}{r|}{44.2}              & 44.5         \\ \hline
\multicolumn{1}{|c|}{MMLU ~\cite{hendrycks2020measuring}}        & \multicolumn{1}{r|}{37.4}          & \multicolumn{1}{r|}{37.9}              & 38.1         \\ \hline
\multicolumn{1}{|c|}{ARC-Easy ~\cite{clark2018think}}            & \multicolumn{1}{r|}{60.5}          & \multicolumn{1}{r|}{61.9}              & 61.4         \\ \hline
\multicolumn{1}{|c|}{WinoGrande ~\cite{sakaguchi2021winogrande}} & \multicolumn{1}{r|}{54.6}          & \multicolumn{1}{r|}{54.5}              & 54.8         \\ \hline
\end{tabular}
    }
\end{minipage}
\end{table}

\subsection{Hash Functions}
\label{sec:hash_function}

\paragraph{\textit{\textbf{Computational performance}}}
We compare the computational performance of our reusable hash functions with MurmurHash3~\cite{appleby2012murmurhash3}, a widely used non-cryptographic hash function adopted in the GPU baseline. To ensure a fair comparison, we implement a CUDA version of MurmurHash3 and measure only the GPU kernel execution time corresponding to the hash computation, excluding other components of the MinHash generation pipeline (e.g., data loading, memory transfers). 
As reported in \autoref{tab:result6a}, across datasets of varying sizes, our hash function consistently outperforms MurmurHash3 in terms of kernel execution speed. On the RealNews dataset, the proposed hash function achieves a 4.85$\times$ speedup over MurmurHash3.

\begin{figure}[!t]
\centering
\includegraphics[width=0.9\linewidth]{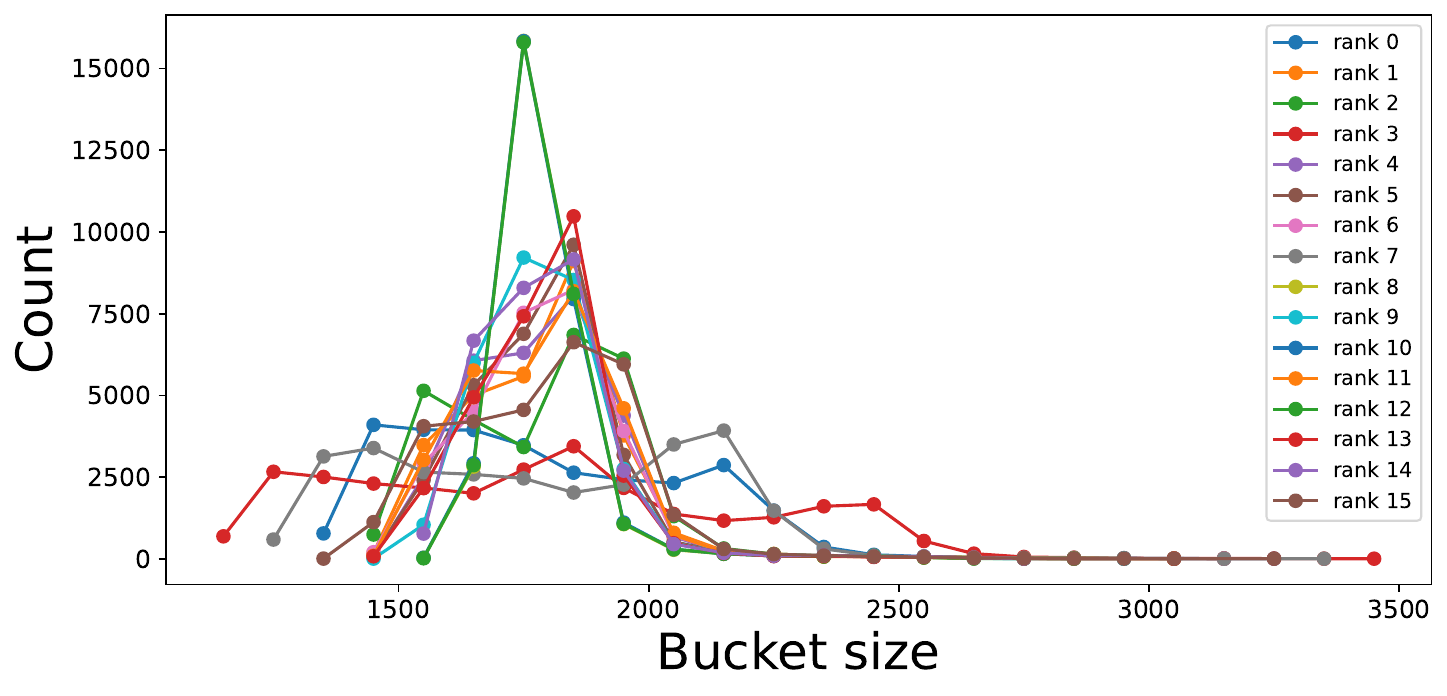}
\caption{Bucket size distribution per rank.}
\label{fig:bucket_dist}
\end{figure}

\paragraph{\textit{\textbf{Imbalance in Buckets}}}
We conducted a thorough analysis of the distribution of documents across buckets in the C4 dataset to assess potential imbalances during our MinHash LSH process. As shown in \autoref{fig:bucket_dist}, the distribution of the number of documents per bucket varies slightly for each rank. However, most buckets cluster around the mean value, which is calculated by dividing the total number of documents by the total number of buckets.

Outlier buckets—defined as those whose sizes exceed the mean by more than twice the standard deviation—account for only about 1.2\% of all buckets. Additionally, the pairwise comparisons within these outliers represent merely 3.6\% of the total per-band comparisons, indicating a negligible impact. The difference between the actual number of pairwise comparisons and the expected number based on a perfectly uniform distribution is approximately 1.97\%, which is not statistically significant.

These results indicate that, while minor imbalances are present, their effect on the overall number of comparisons is minimal. Thus, the distribution can be considered effectively uniform in practice. Additionally, the number of pairwise comparisons conducted per rank is nearly constant, further supporting the notion of a balanced workload distribution across ranks. Detailed experimental results can be found in the Appendix ~\ref{app:imbalance}.

\begin{figure*}[!t]
  \centering
  \begin{minipage}{0.48\linewidth}
    \centering
    \includegraphics[width=0.9\linewidth]{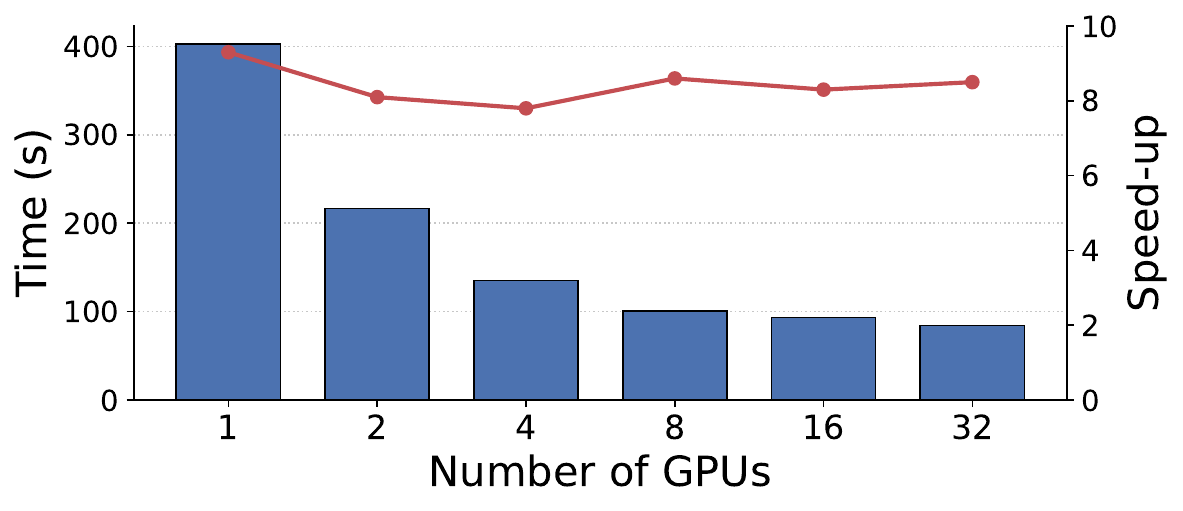}
    \caption*{(a) Result on the RealNews dataset.}
  \end{minipage}
  \hfill
  \begin{minipage}{0.48\linewidth}
    \centering
    \includegraphics[width=0.9\linewidth]{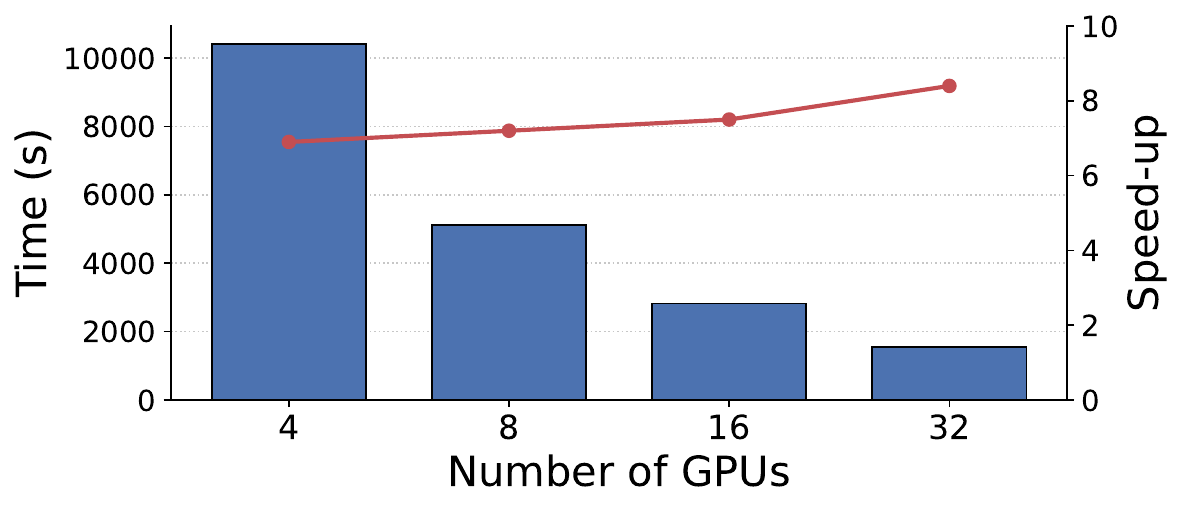}
    \caption*{(b) Result on the full C4 dataset.} 
  \end{minipage}
  \caption{Scalability with the number of GPUs. Blue bars represent the execution time of \sys, while the red line indicates the speedup over the GPU baseline.}
  \label{fig:multi_gpu}
\end{figure*}

\begin{figure}[!ht]
\centering
\includegraphics[width=\linewidth]{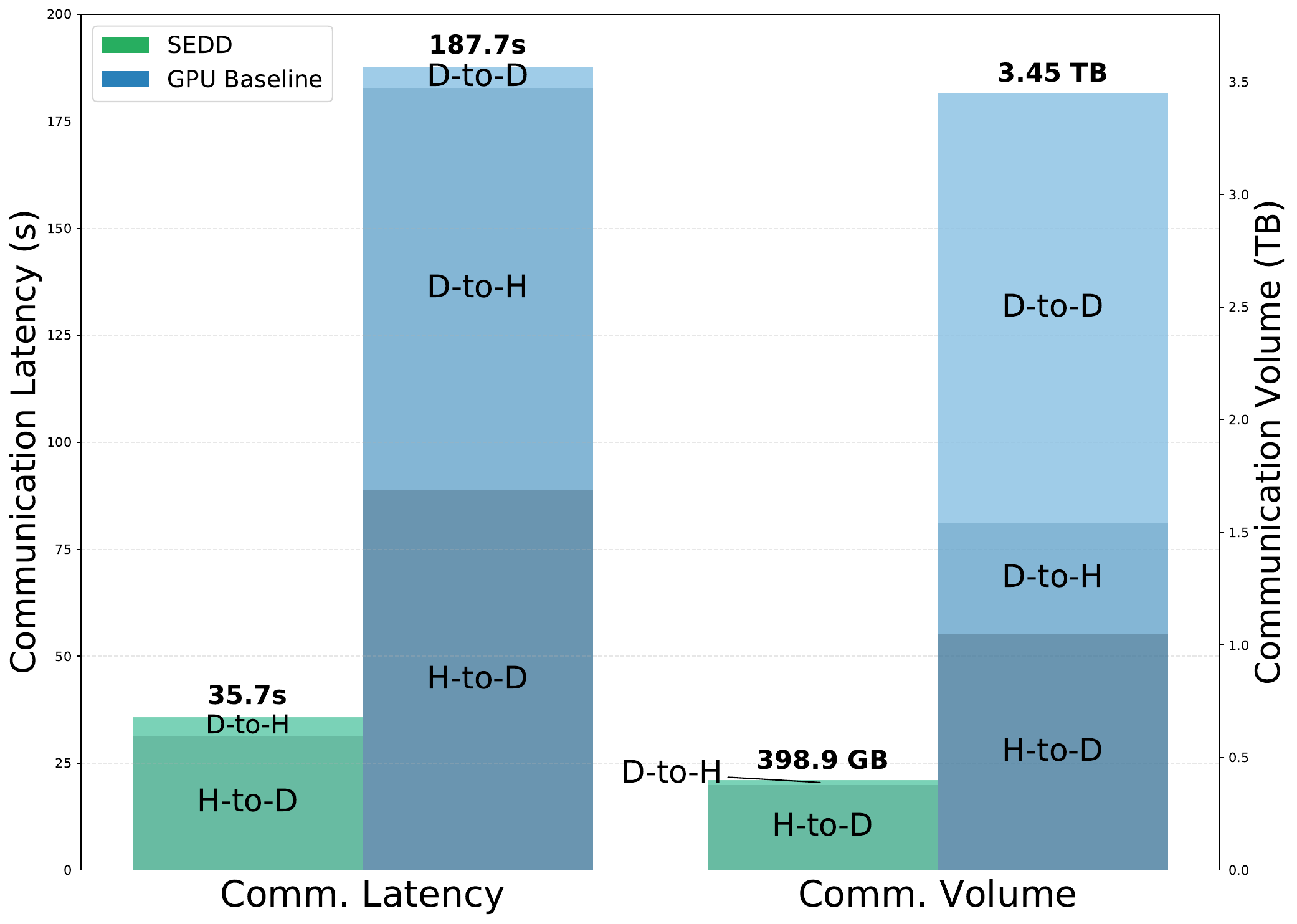}
\caption{Comparison of communication latency and volume between \sys and the GPU baseline on the RealNews dataset using 4 GPUs.}
\label{fig:comm_comparison}
\end{figure}

\subsection{Scalability}
\label{sec:scalability}
By varying the number of GPUs, the size of the dataset, and the underlying hardware architecture, we evaluate the comprehensive scalability and robustness of \sys.

\paragraph{\textit{\textbf{Number of GPUs}}}
As illustrated in \autoref{fig:multi_gpu}, we evaluate the execution time and speedup as the number of GPUs increases from 1 to 32. Configurations with eight or more GPUs correspond to a multi-node setup. 
For the RealNews dataset, \sys achieves a 50$\times$ speedup over the CPU baseline even with a single GPU, and completes the process in under 90 seconds when using 32 GPUs. Throughout the scaling process, \sys consistently maintains at least a 7$\times$ speedup over the GPU baseline.
For the full C4 dataset, execution time decreases nearly linearly as the number of GPUs increases. In particular, deduplication of 365M documents is completed in approximately 25 minutes using 32 GPUs. Interestingly, the speedup over the GPU baseline increases as we transition to a multi-node environment, indicating that \sys’s streaming extraction and communication–computation overlapping effectively mitigate synchronization and I/O overheads that typically limit distributed LSH pipelines.
We also observe that scaling efficiency depends strongly on dataset size. For relatively smaller datasets such as RealNews, the workload per GPU becomes insufficient as more devices are added, and fixed communication and synchronization overheads begin to dominate, leading to performance saturation beyond eight GPUs. In contrast, larger datasets such as C4 provide sufficient computational intensity to fully utilize additional hardware resources, allowing \sys to sustain strong scaling and higher speedups in large-scale cluster environments.

\paragraph{\textit{\textbf{Dataset size.}}}
To evaluate whether \sys can process trillion-scale corpora at practical speeds, we measure its performance using 32 GPUs across datasets of increasing size: RealNews, the full C4 dataset, and RedPajama-1T~\cite{together2023redpajama}. 
For the relatively small RealNews dataset, \sys completes deduplication in under 90 seconds, confirming its high throughput even at modest scales. When applied to the full C4 dataset (approximately 156 billion tokens), the total processing time is 1,545.5 seconds (about 25.8 minutes), with 369.1 seconds spent on signature generation and 1,144.8 seconds on pairwise comparison.
Most notably, \sys processes the 1.2 trillion tokens of the RedPajama-1T corpus in 10,458.1 seconds (approximately 2.9 hours). In this trillion-token setting, signature generation requires 1,401.4 seconds, while the exhaustive pairwise comparison phase takes 9,056.7 seconds. These results demonstrate that \sys sustains high throughput as dataset scale increases, keeping the total processing time within practical limits for modern LLM data pipelines. Given that state-of-the-art LLM training increasingly relies on trillion-token corpora, \sys provides a scalable and practical solution for large-scale data deduplication.

\paragraph{\textit{\textbf{Robustness across GPU Architectures.}}}
To verify the portability of \sys, we evaluate its performance across different GPU architectures beyond our V100 cluster. We conducted additional tests on a workstation equipped with NVIDIA RTX 3090 GPUs. Despite the architectural differences, \sys successfully deduplicated the RealNews dataset in 119 seconds using 4 GPUs. This confirms that our hardware-aware parameter selection and optimized CUDA kernels effectively adapt to varying hardware specifications.

\subsection{Analysis of Communication Efficiency}

\label{sec:comm_analysis}
To further investigate the architectural advantages of \sys, we conduct a comparative analysis of communication latency and volume during the deduplication of the RealNews dataset in a 4-GPU environment. \autoref{fig:comm_comparison} illustrates the breakdown of data transfers into three categories: Host-to-Device (H-to-D), Device-to-Host (D-to-H), and Device-to-Device (D-to-D).

As shown in the results, the GPU baseline suffers from substantial communication overhead, with a total latency of 187.7 seconds and a communication volume reaching 3.45 TB. This inefficiency is primarily driven by the physical shuffling phase, which requires extensive D-to-D and D-to-H data movements to reorganize signatures into their respective buckets across distributed GPUs. In contrast, \sys demonstrates significantly higher efficiency, reducing the communication latency to 35.7 seconds (a 5.2x speedup) and the total volume to 398.9 GB (an 8.6x reduction).

The reduction in communication volume is particularly notable; by utilizing a streaming extraction strategy, \sys eliminates the need for large-scale D-to-D shuffling. Instead, it only performs the necessary H-to-D and D-to-H transfers for signature processing. Furthermore, \sys effectively mitigates the remaining communication latency through the communication-computation overlapping techniques described in Section~\ref{sec:opt}. By pipelining data transfers with GPU kernel execution, \sys hides the majority of the data movement overhead, ensuring that communication does not become a bottleneck as the dataset scale increases.
\section{Conclusion}
We present a framework called \sys, which performs MinHash LSH-based deduplication efficiently on GPUs. \sys significantly outperforms the CPU baseline included in SlimPajama by up to 158$\times$ and the GPU baseline included in NVIDIA NeMo Curator by up to 7.8$\times$ on a single node with four GPUs when processing a dataset with 30M documents. In a multi-node and multi-GPU environment, \sys completes deduplication of 1.2 trillion tokens in approximately 3 hours.
These performance gains are enabled by a series of GPU-oriented system optimizations, including reusable rolling hash functions for efficient MinHash generation and a streaming execution pipeline that overlaps communication and computation. As a result, \sys achieves both high throughput and high accuracy: documents identified as duplicates exhibit a Jaccard similarity of 0.95 or higher compared to those found by the standard MinHash algorithm. Extensive experimental results demonstrate that \sys is a practical and scalable solution for deduplication in the large-scale datasets used for modern LLM training.



\section*{Acknowledgement}
This work was partially supported by the National Research Foundation of Korea (NRF) under Grant No. RS-2023-00222663 (Center for Optimizing Hyperscale AI Models and Platforms), and by the Institute for Information and Communications Technology Promotion (IITP) under Grant No. 2018-0-00581 (CUDA Programming Environment for FPGA Clusters) and No. RS-2025-02304554 (Efficient and Scalable Framework for AI Heterogeneous Cluster Systems), all funded by the Ministry of Science and ICT (MSIT) of Korea. It was also partially supported by the Korea Health Industry Development Institute (KHIDI) under Grant No. RS-2025-25454559 (Frailty Risk Assessment and Intervention Leveraging Multimodal Intelligence for Networked Deployment in Community Care), funded by the Ministry of Health and Welfare (MOHW) of Korea. Additional support was provided by the BK21 Plus Program for Innovative Data Science Talent Education (Department of Data Science, Seoul National University, No. 5199990914569) and the BK21 FOUR Program for Intelligent Computing (Department of Computer Science and Engineering, Seoul National University, No. 4199990214639), both funded by the Ministry of Education (MOE) of Korea. This work was also partially supported by the Artificial Intelligence Industrial Convergence Cluster Development Project, funded by the MSIT and Gwangju Metropolitan City. Research facilities were provided by the Institute of Computer Technology (ICT) at Seoul National University.

\newpage
\bibliographystyle{ACM-Reference-Format}

\newpage
\appendix
\section*{\centering\textbf{Appendix}}


\section{MinHash}
\label{app:MinHashGen}
\subsection{Jaccard Similarity}

Jaccard similarity~\citep{jaccard:1912}, {\small $J(A, B)$}, is a metric to quantify the similarity of two finite sets, {\small $A$} and {\small $B$}:
\begin{equation}
J(A, B) = \frac{|A \cap B|}{|A \cup B|} 
\label{eq:jaccard}
\end{equation}
When used as a metric for finding document similarity, it is defined as the number of common words between the documents divided by the total number of words in them. While this is the most intuitive way to measure similarity, calculating the Jaccard similarity directly between documents in a large corpus is not feasible due to the computational cost. Thus, the MinHash algorithm~\cite{broder1997resemblance}, which approximates Jaccard similarity, is often used instead.

\begin{figure}[!b]
\centering
\includegraphics[width=0.7\linewidth]{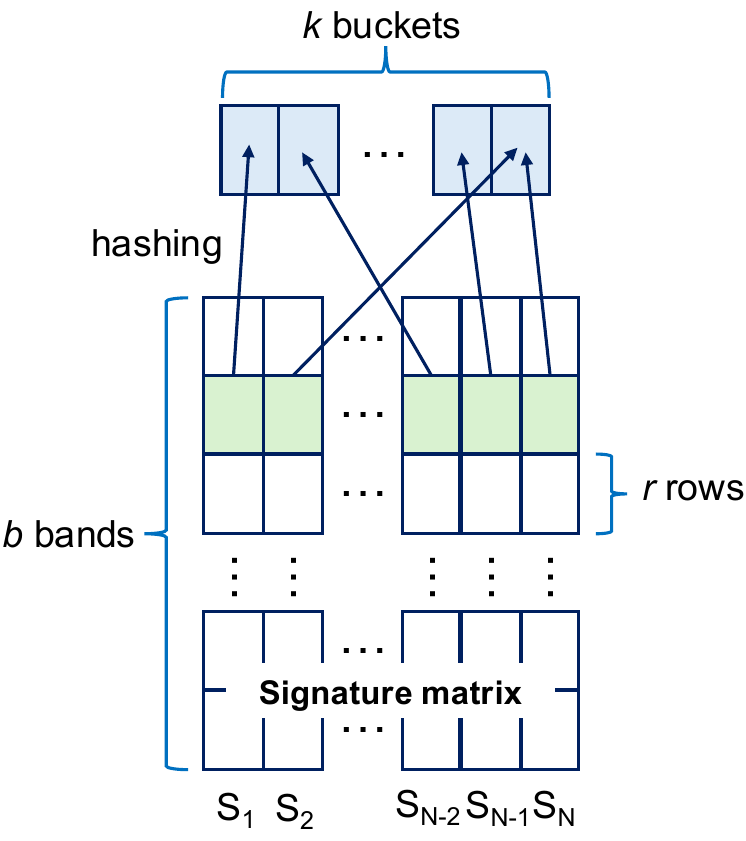}
\caption{Hashing by MinHash LSH.}
\label{fig:lsh}
\end{figure}

\subsection{MinHash LSH}
\label{app:minhashlsh}
The critical difference between MinHash and MinHash LSH lies in the stage of generating duplicate pairs. In MinHash LSH, the signature column vector in the signature matrix of each document is divided into $b$ \textit{bands}, each of which has $r$ integers. As shown in \autoref{fig:lsh}, LSH hashes the bands in each document to $k$ buckets. Let $S_i$ be the signature vector of the document $D_i$. For a given signature vector $S_1$ and $S_2$, when at least a pair of bands $B_{S_1}$ from $S_1$ and $B_{S_2}$ from $S_2$ hashed to the same bucket, we tag them as candidate pairs, potentially similar documents. For example, in \autoref{fig:lsh}, $D_2$ and $D_N$ are potentially similar documents because they have bands hashed to the same bucket. Then, LSH performs pairwise comparisons on each document's signature vectors of length $b \times r$  for the candidate pairs. If the similarity between the two vectors exceeds a predefined threshold, the two documents are considered a duplicate pair. Once the union graph is constructed for each band, the union graphs for all bands are merged to obtain the final union graph.



\section{Suitability of \sys's Hash Functions}
\label{app:suit_hash}
In general, a good hash function for MinHash should have the following properties:
\begin{itemize}
    \item \textbf{Determinacy}: The same input produces the same output.
    \item \textbf{Uniformity}: The hash function should distribute outputs uniformly across the range.
    \item \textbf{Collision resistance}: Two distinct inputs should unlikely produce the same hash value.
\end{itemize}
The proposed hash function satisfies these properties. Determinacy and uniformity are straightforward. Collision resistance is also met with a large value of $p$. When uniformity is satisfied, the probability of a collision approaches $\frac{1}{p}$, which is very small. This follows the same intuition as the Schwartz–Zippel lemma~\cite{schwartz1980}. This ensures collisions are rare when $p$ is sufficiently large. 

In many contexts, hash functions are required to resist preimage attacks, making it computationally difficult to reverse-engineer the input from the hash value. However, it is not a concern in MinHash because the hash function is not used for security purposes. Moreover, while other non-cryptographic hash functions such as FarmHash~\cite{farmhashBlog} and xxHash~\cite{xxhashRepo} exist, our design provides the \textit{rolling property}, which allows efficient reuse of previously computed results when sliding over substrings. This property yields better throughput for string hashing compared to FarmHash or xxHash.




\section{Hardware-aware Parameter Selection}
\label{app:auto_param_selection}

\sys employs an automated, hardware-aware mechanism to determine execution parameters, aiming to maximize throughput while maintaining stable operation across diverse hardware environments.

\paragraph{\textit{\textbf{Adaptive Storage Strategy}}}
The system first determines whether intermediate results—specifically the signature matrix and bucket IDs—should be keeped in main memory or offloaded to secondary storage. Let $N$ denote the total number of documents, $H$ the signature length, and $n_{\text{proc}}$ the number of CPU processes per node. The required memory for intermediate buffers is estimated as
\[
M_{req} = N \times (H + 1) \times \text{sizeof(int)} \times n_{\text{proc}}.
\]

This estimate is compared against the available host memory $M_{avail\_cpu}$. If
\[
M_{req} \le \alpha \cdot M_{\text{avail\_cpu}},
\]
the system operates fully in memory to eliminate disk I/O overhead. Otherwise, \sys automatically activates a disk-based offloading strategy and persists intermediate hash results in Parquet format to prevent memory exhaustion when scaling to trillion-token datasets. We use a conservative safety margin of $\alpha = 0.2$ to account for transient memory usage from the operating system and runtime buffers.

\paragraph{\textit{\textbf{Dynamic Pipelining with Parameter \ $C$}}}
During the streaming extraction phase, each process handles $C$ bucket IDs concurrently to reduce the number of full-dataset scans. The value of $C$ is dynamically tuned to utilize available CPU memory without triggering swap overhead. Given $n_{\text{proc}}$ concurrent processes per node and $K$ total buckets, \sys determines $C$ using the constraint
\begin{equation}
\label{eq:C_param}
C \cdot n_{\text{proc}} \cdot \left( \frac{N}{K} \right) \cdot H \cdot \text{sizeof(int)} \le \alpha \cdot M_{\text{avail\_cpu}},
\end{equation}
where $M_{\text{avail\_cpu}}$ denotes the currently available memory and $\alpha$ is the safety margin factor. Maximizing $C$ within this bound reduces redundant file accesses and helps hide I/O latency, as more buckets can be verified in a single pass over the stored hash results.

\paragraph{\textit{\textbf{GPU Memory Constraint Management}}}
Finally, the system enforces a hard limit on the maximum bucket size $D_{\max}$ to ensure compatibility with the available GPU memory $M_{gpu}$. Since the comparison kernel performs exhaustive $O(D^2)$ pairwise checks, the required memory for processing a bucket depends on both the number of documents $D$ and the signature length $H$. \sys ensures that
\[
D \times H \times \text{sizeof(int)} < \beta \cdot M_{gpu}.
\]
where $\beta$ is the safety margin factor. If a bucket exceeds $D_{\max}$, it is partitioned into smaller sub-units to maintain memory safety and prevent Out-of-Memory (OOM) errors during kernel execution.

\begin{table}[!t]
\centering
\begin{minipage}{0.9\linewidth}
    \centering
\caption{Ablation study on the RealNews dataset using 4 GPUs.}
\label{tab:ablation_results}
    \vspace{-0.5\baselineskip}
    \resizebox{0.9\linewidth}{!}{%
\begin{tabular}{lc}
\hline
Configuration & Execution Time (s) \\ \hline
\sys (Full Optimization) & 135.2 \\
w/ MurmurHash3  & 147.7 \\
w/o Double Buffering (Sequential I/O) & 189.0 \\ \hline
\end{tabular}
    }
\end{minipage}
  \vspace{0.3cm} 

\begin{minipage}{0.9\linewidth}
    \centering
    \caption{File read and buffering performance with varying the value of the parameter $C$ for the RealNews dataset (in seconds).}
    \label{tab:result6b}
    \vspace{-0.5\baselineskip}
    \resizebox{0.9\linewidth}{!}{%
    \begin{tabular}{|c||r|r|r|r|r|}
    \hline
    $C$ & 256 & 512 & 1,024 & 2,048 & 4,096 \\ \hline \hline
    File read & 1,168.5 & 580.2 & 160.5 & 75.8 & 32.6 \\ \hline
    Buffering &  125.7 & 78.2 & 28.7 & 22.1 & 18.1  \\ \hline
    \begin{tabular}{c}
File read \&   \\
Buffering
\end{tabular}  & 1,294.2 & 658.4 & 189.2 & 97.9 & 50.7 \\ \hline
    \end{tabular}
    }
\end{minipage}
  \vspace{0.3cm} 

\begin{minipage}{\linewidth}
    \centering
    \caption{Comparison time in seconds by varying the value of the parameter $K$ for the RealNews dataset.}
    \label{tab:result6c}
    \vspace{-0.5\baselineskip}
    \resizebox{0.99\linewidth}{!}{%
\begin{tabular}{|c||r|r|r|r|}
\hline
$k$ & 2 & 4 & 6 & 8 \\ \hline \hline
\# of buckets & 11,457 & 22,915 & 34,373 & 45,380 \\ \hline
Comparison time (s) & 118.5 & 86.2 & 88.1 & 94.4 \\ \hline
\# of Removed docs& 3,130,190 & 3,130,191 & 3,130,191 & 3,130,191 \\ \hline
\end{tabular}

    }
\end{minipage}
\end{table}

\section{Ablation on System-Level Design Choices}
\label{app:ablation}

To better understand the contribution of each component in \sys, we conduct a series of ablation studies on key system-level design choices. In particular, we analyze how different architectural optimizations and parameter settings affect the overall performance of MinHash LSH-based deduplication.

\paragraph{\textit{\textbf{Ablation study}}}
We provide a quantitative breakdown of the performance gains contributed by each architectural component of \sys. Table~\ref{tab:ablation_results} summarizes the execution time for deduplicating the RealNews dataset using 4 GPUs under different system configurations.

The results show that the fully optimized version of \sys achieves the best performance, completing the pipeline in 135.2 seconds. Notably, even when using the standard MurmurHash3, our framework executes in 147.7 seconds, which is already substantially faster than traditional baselines due to our hand-tuned GPU kernel implementation. Replacing MurmurHash3 with our proposed rolling hash further accelerates the hash generation stage, reducing the overall execution time by an additional 12.5 seconds. 

Furthermore, disabling the double buffering mechanism significantly degrades performance, increasing the execution time to 189.0 seconds. This result highlights the importance of overlapping storage I/O with GPU computation in achieving high throughput.

\paragraph{\textit{\textbf{The effect of $\boldsymbol{C}$}}}
$C$ denotes the number of buckets processed by each process in a single step, meaning that a larger $C$ allows more buckets to be handled per file I/O operation. While we set $C$ to a practical value considering CPU memory constraints, we further analyze its impact on deduplication performance.

\autoref{tab:result6b} reports the file I/O time measured during the comparison stage on the RealNews dataset while varying $C$. The results show that increasing $C$ consistently reduces the file I/O overhead, demonstrating that $C$ plays a crucial role in improving throughput. Therefore, in practice, we set $C$ to the largest value permitted by available system resources to minimize file I/O cost.

\paragraph{\textit{\textbf{The effect of $\boldsymbol{K}$}}}
As discussed earlier, the optimal number of buckets can be expressed as $K = kN^{1/2}$. We evaluate the deduplication performance on the RealNews dataset using different values of $k$ and describe the process for selecting an appropriate configuration. As shown in \autoref{tab:result6c}, increasing $k$ beyond a certain threshold does not increase the number of deduplicated documents, while the comparison time begins to plateau. Based on experiments across multiple datasets, we find that setting $k = 4$ provides the best trade-off between computational cost and deduplication effectiveness for \sys.

\section{System Configuration}
\label{app:system_config}
We use a 8-node GPU cluster with a storage node for our experiments. Each node is equipped with
four NVIDIA Tesla V100 GPUs, with each GPU having 32GB of memory. \autoref{tab:system_conf} provides the detailed target system configuration.

\begin{table}[h]
\centering
\caption{System configuration of the 8-node GPU cluster}
\label{tab:system_conf}
\vspace{-0.5\baselineskip}
\resizebox{\linewidth}{!}{%
\begin{tabular}{|cl|}
\hline\hline
\multicolumn{2}{|c|}{\textbf{GPU node}}                                                                  \\ \hline\hline
\multicolumn{1}{|c|}{CPU}               & 1 $\times$ AMD EPYC 7502 32-Core Processor                \\ \hline
\multicolumn{1}{|c|}{Memory}            & 8 $\times$ 64GB DDR4 DIMM                                \\ \hline
\multicolumn{1}{|c|}{GPU}               & 4 $\times$ NVIDIA Tesla V100                             \\ \hline
\multicolumn{1}{|c|}{OS}                & Ubuntu 20.04.6 (kernel 5.4.0-100)                         \\ \hline
\multicolumn{1}{|c|}{Compiler}          & nvcc 12.4, GCC 9.4                                        \\ \hline
\multicolumn{1}{|c|}{GPU Driver}        & 520.61.05                                                 \\ \hline
\multicolumn{1}{|c|}{MPI Version}       & 4.1                                                       \\ \hline
\multicolumn{1}{|c|}{CUDA Version}      & 12.4                                                      \\ \hline
\multicolumn{2}{|c|}{\textbf{Storage node}}                                                                    \\ \hline\hline
\multicolumn{1}{|c|}{CPU}               & 
\begin{tabular}{l}
2 $\times$ AMD EPYC 7502 32-Core Processor 
\end{tabular}\\ \hline
\multicolumn{1}{|c|}{Memory}            & 
\begin{tabular}{l}
8 $\times$ 64GB DDR4 DIMM  
\end{tabular}
\\ \hline

\multicolumn{1}{|c|}{
\begin{tabular}{c}
RAID \\
Configuration
\end{tabular}
}            & 
\begin{tabular}{l}
RAID5, \\
7 $\times$ 3.7TB Sabrent Rocket 4.0 Plus NVMe SSDs, \\
\end{tabular}
\\ \hline
\multicolumn{1}{|c|}{File System}        & 
\begin{tabular}{l}
ext4        
\end{tabular}\\ \hline
\multicolumn{1}{|c|}{Network Interface} & 
\begin{tabular}{l}
InfiniBand  (200Gb/s)   
\end{tabular}\\ \hline
\end{tabular}
}
\end{table}

\section{Language model configuration}
\label{app:model_config}
The detailed configuration of the LLaMA 1B model used in the language model experiments described in Section~\ref{sec:accuracy} is provided in \autoref{tab:model_config}.

\begin{table}[!h]
\centering
\caption{Model configuration.}
\label{tab:model_config}
\begin{minipage}{\linewidth}
    \centering
    \vspace{-0.5\baselineskip}
    \resizebox{0.9\linewidth}{!}{%
    \begin{tabular}{|l|l|}
    \hline
    \textbf{Model Architecture} & LLaMA 3.2 \\ \hline
    \textbf{Number of Layers} & 16 \\ \hline
    \textbf{Hidden Size} & 2048 \\ \hline
    \textbf{Intermediate Size} & 8192 \\ \hline
    \textbf{Attention Heads} & 32 \\ \hline
    \textbf{Sequence Length} & 2048 \\ \hline
    \textbf{Batch Size} & 1M tokens \\ \hline
    \textbf{Learning Rate Scheduler} & Cosine Decay, Warmup steps=10\% \\ \hline
    \textbf{Optimizer} & AdamW \\ \hline
    \textbf{Learning Rate} & 3e-4 \\ \hline
    \end{tabular}
    }
\end{minipage}

\end{table}

\begin{table*}[h]
\centering
\caption{Per-rank statistics of bucket sizes and pairwise comparisons in the C4 dataset.}
\label{tab:sup_outlier}
\begin{tabular}{r|rrrrrrr}
\toprule
\textbf{Rank} & 
\textbf{Mean} & 
\textbf{StdDev} & 
\textbf{Comparisons} & 
\textbf{Outliers} & 
\textbf{\%Out} & 
\textbf{OutComp} & 
\textbf{\%OutComp} \\
\midrule
0  & 1,790.82 & 331.52 & 47,516,269,551 & 264 & 0.92 & 3,091,775,767 & 3.25 \\
1  & 1,790.82 & 238.53 & 46,756,449,092 & 387 & 1.35 & 3,594,688,373 & 3.84 \\
2  & 1,790.82 & 248.64 & 46,827,051,391 & 373 & 1.30 & 3,408,740,924 & 3.64 \\
3  & 1,790.82 & 425.97 & 47,541,685,437 & 243 & 0.85 & 2,933,144,397 & 3.02 \\
4  & 1,790.82 & 215.24 & 46,604,982,146 & 380 & 1.33 & 3,167,172,510 & 3.40 \\
5  & 1,790.82 & 259.04 & 46,902,741,222 & 310 & 1.08 & 3,558,005,800 & 3.79 \\
6  & 1,790.82 & 232.75 & 46,717,393,506 & 371 & 1.29 & 3,397,637,751 & 3.63 \\
7  & 1,790.82 & 331.54 & 47,516,420,018 & 278 & 0.97 & 2,487,274,545 & 2.62 \\
8  & 1,790.82 & 147.99 & 46,254,877,288 & 556 & 1.94 & 3,504,375,831 & 3.79 \\
9  & 1,790.82 & 210.98 & 46,578,954,192 & 415 & 1.45 & 3,480,028,218 & 3.73 \\
10 & 1,790.82 & 196.33 & 46,493,455,190 & 387 & 1.35 & 3,247,192,167 & 3.49 \\
11 & 1,790.82 & 282.27 & 47,082,957,868 & 224 & 0.78 & 3,363,723,606 & 3.57 \\
12 & 1,790.82 & 181.62 & 46,413,733,631 & 419 & 1.46 & 3,197,735,770 & 3.44 \\
13 & 1,790.82 & 334.38 & 47,543,508,465 & 169 & 0.59 & 4,151,084,255 & 4.36 \\
14 & 1,790.82 & 193.71 & 46,478,801,548 & 457 & 1.59 & 3,434,621,567 & 3.69 \\
15 & 1,790.82 & 210.72 & 46,577,389,821 & 463 & 1.62 & 3,278,460,404 & 3.52 \\

\bottomrule
\end{tabular}%

\end{table*}

\begin{table*}[]
\centering
\caption{Mean-Only Assumption vs Actual Comparisons Per Rank}
\label{tab:sup_comparison}
\begin{tabular}{c | c c c c}
\toprule
Rank & 
\begin{tabular}{c}Expected \\ Comparisons\end{tabular} & 
\begin{tabular}{c}Actual \\ Comparisons\end{tabular} & 
\begin{tabular}{c}Delta \\ (A-E)\end{tabular} & 
\begin{tabular}{c}RelErr \\ \%\end{tabular} \\
\midrule
0  & 45,940,962,914 & 47,516,269,551 & 1,575,306,637 & 3.32 \\
1  & 45,940,962,914 & 46,756,449,092 &   815,486,178 & 1.74 \\
2  & 45,940,962,914 & 46,827,051,391 &   886,088,477 & 1.89 \\
3  & 45,940,962,914 & 47,541,685,437 & 1,600,722,523 & 3.37 \\
4  & 45,940,962,914 & 46,604,982,146 &   664,019,232 & 1.42 \\
5  & 45,940,962,914 & 46,902,741,222 &   961,778,308 & 2.05 \\
6  & 45,940,962,914 & 46,717,393,506 &   776,430,592 & 1.66 \\
7  & 45,940,962,914 & 47,516,420,018 & 1,575,457,104 & 3.32 \\
8  & 45,940,962,914 & 46,254,877,288 &   313,914,374 & 0.68 \\
9  & 45,940,962,914 & 46,578,954,192 &   637,991,278 & 1.37 \\
10 & 45,940,962,914 & 46,493,455,190 &   552,492,276 & 1.19 \\
11 & 45,940,962,914 & 47,082,957,868 & 1,141,994,954 & 2.43 \\
12 & 45,940,962,914 & 46,413,733,631 &   472,770,717 & 1.02 \\
13 & 45,940,962,914 & 47,543,508,465 & 1,602,545,551 & 3.37 \\
14 & 45,940,962,914 & 46,478,801,548 &   537,838,634 & 1.16 \\
15 & 45,940,962,914 & 46,577,389,821 &   636,426,907 & 1.37 \\

\midrule
\multicolumn{1}{c}{\textbf{Total}} & 735,055,406,624 & 749,806,670,366 & 14,751,263,742 & 1.97 \\
\bottomrule
\end{tabular}

\end{table*}

\section{Load imbalance analysis}
\label{app:imbalance}
\autoref{tab:sup_outlier} presents the per-rank statistics of bucket sizes and outlier comparisons in the C4 dataset. 
Each row corresponds to a rank (i.e., a band in the MinHash LSH process), and the columns are defined as follows:

\begin{itemize}[leftmargin=*, itemsep=2pt]
    \item \textbf{Mean}: the average number of documents per bucket for the given rank.
    \item \textbf{StdDev}: the standard deviation of the number of documents across buckets, indicating the spread of bucket sizes.
    \item \textbf{Comparisons}: the total number of pairwise comparisons for the rank.
    \item \textbf{Outliers}: the number of buckets whose size exceeds the mean by more than two standard deviations.
    \item \textbf{\%Out}: the percentage of buckets that are considered outliers relative to the total number of buckets.
    \item \textbf{OutComp}: the total number of pairwise comparisons performed within outlier buckets.
    \item \textbf{\%OutComp}: the proportion of comparisons in outlier buckets relative to the total number of per-rank comparisons.
\end{itemize}

The number of pairwise comparisons reported in \autoref{tab:sup_outlier} is computed based on the actual number of documents in each bucket during the MinHash LSH process on the C4 dataset.
For a bucket containing $n$ documents, the total number of pairwise comparisons is calculated using the formula $n(n-1)/2$, which corresponds to summing all pairs of documents within the bucket. 
Similarly, the \textbf{OutComp} column accounts for comparisons in outlier buckets (buckets whose size exceeds the mean by more than two standard deviations) using the same $n(n-1)/2$ calculation per bucket. 
Outlier buckets—defined as those whose size exceeds the mean by more than two standard deviations—constituted a small fraction of all buckets, typically around 0.59\%--1.94\% per rank. 
The number of pairwise comparisons within these outlier buckets (\textbf{OutComp}) accounted for only approximately 2.62\%--4.36\% of the total per-rank comparisons, indicating that the contribution of outliers to the overall computation is minimal and can be considered negligible.

\autoref{tab:sup_comparison} reports the comparison counts per rank under the MinHash LSH process, illustrating the difference between a hypothetical uniform distribution and the actual computation. 
Each column is defined as follows:

\begin{itemize}[leftmargin=*, itemsep=2pt]
    \item \textbf{Expected Comparisons}: the total number of pairwise comparisons assuming that all buckets contain exactly the same number of documents.
    \item \textbf{Actual Comparisons}: the total number of pairwise comparisons actually performed for the rank.
    \item \textbf{Delta (A-E)}: the difference between the actual and expected comparisons.
    \item \textbf{RelErr \%}: the relative error expressed as a percentage of the expected comparisons.
\end{itemize}

As shown in \autoref{tab:sup_comparison}, while individual ranks exhibit small deviations from the expected counts, the total number of actual comparisons across all ranks differs from the perfectly uniform case by only approximately 1.97\%. 
This indicates that although minor imbalances exist in document distribution among buckets, their overall impact on the total number of pairwise comparisons is minimal.

\end{document}